\pdfoutput=1

\documentclass[11pt]{article}

\usepackage[]{acl}

\usepackage{times}
\usepackage{latexsym}
\usepackage{pgfplots}
\usepackage[T1]{fontenc}
\usepackage{mathrsfs}
\usepackage{algorithm}
\usepackage{algorithmic}
\usepackage{bm}
\usepackage{amsmath}
\usepackage{cleveref}
\newcommand{\softmax}{\mathrm{softmax}}
\crefname{section}{§}{§§}

\usepackage[utf8]{inputenc}

\usepackage{microtype}
\makeatletter
\newcommand{\printfnsymbol}[1]{%
  \textsuperscript{\@fnsymbol{#1}}%
}
\makeatother
\title{Reasoning over Hybrid Chain for Table-and-Text Open Domain QA}

\author{Wanjun Zhong$^{1\dag}$\thanks{\ \ \ Indicates equal contribution}, Junjie Huang$^{3*}$\thanks{\ \ \ Work done while this author was an intern at Microsoft Research.}, Qian Liu$^{3\dag}$, Ming Zhou$^4$,\\ \textbf{Jiahai Wang$^1$, Jian Yin$^1$ and Nan Duan$^2$} \\
	$^1$ The School of Computer Science and Engineering, Sun Yat-sen University.\\
	$^2$ Microsoft Research \quad $^3$ Beihang University \quad $^4$ Langboat Technology\\
	{\tt \{zhongwj25@mail2, wangjiah@mail,issjyin@mail\}.sysu.edu.cn}\\
	{\tt \{huangjunjie, qian.liu\}@buaa.edu.cn} \\
	{\tt nanduan@microsoft.com}; \tt zhouming@chuangxin.com\\ 
}

\begin{document}
	\maketitle
	\begin{abstract}
Tabular and textual question answering requires systems to perform reasoning over heterogeneous information, considering table structure, and the connections among table and text. In this paper, we propose a ChAin-centric Reasoning and Pre-training framework (CARP). CARP utilizes hybrid chain to model the explicit intermediate reasoning process across table and text for question answering. We also propose a novel chain-centric pre-training method, to enhance the pre-trained model in identifying the cross-modality reasoning process and alleviating the data sparsity problem. This method constructs the large-scale reasoning corpus by synthesizing pseudo heterogeneous reasoning paths from Wikipedia and generating corresponding questions. We evaluate our system on OTT-QA, a large-scale table-and-text open-domain question answering benchmark, and our system achieves the state-of-the-art performance. Further analyses illustrate that the explicit hybrid chain offers substantial performance improvement and interpretablity of the intermediate reasoning process, and the chain-centric pre-training boosts the performance on the chain extraction.

	\end{abstract}
	\section{Introduction}
	\begin{figure}[t]
	\vspace{3mm}
		\centering
		\includegraphics[width=.45\textwidth]{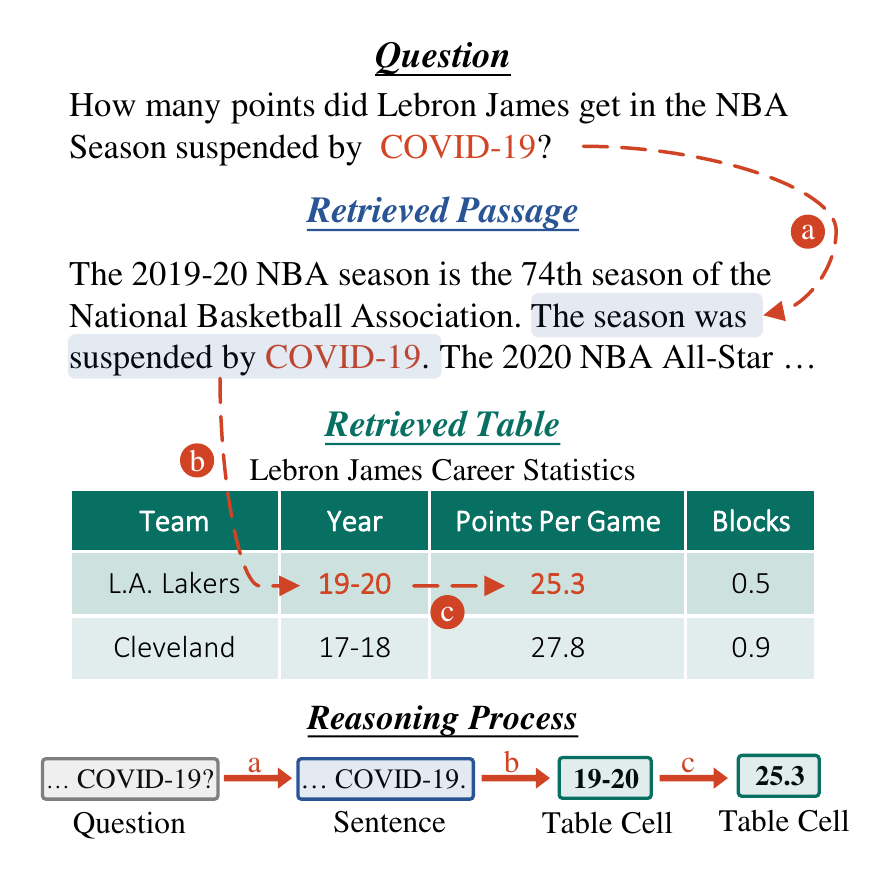}
		\caption{An example of the table-and-text QA with intermediate reasoning process. The answer is $25.3$.}
		\label{fig:example}
	\vspace{-3mm}
	\end{figure}
	Open domain question answering \cite{joshi2017triviaqa,dunn2017searchqa,nqopen} requires systems to retrieve and perform reasoning over supported knowledge, and finally derive an answer. 
	Generally, the real-world knowledge resource is heterogeneous, which involve both semi-structured web tables and unstructured text like Wikipedia passages.
	Therefore, question answering over hybrid tabular and textual knowledge is essential and attracts wide attentions \cite{chen2020open}, and is more challenging as systems need to aggregate information in both table and text considering their connections and the table structure.  
	
	As the example shown in Fig. \ref{fig:example}, the complete reasoning process for answering the question involves hybrid information pieces in both the table (``\textit{Year}" and ``\textit{Points}" columns in the first row) and the passage (``\textit{COVID-19}"). 
	Therefore, modeling the structural connections inside heterogeneous knowledge is critical for modeling the reasoning process. 
	Many recent works on table-and-text open domain QA simply take the supported flattened table and passages \cite{chen2020open,li2021dual} as a whole for question answering, which neglects the structural information and connections among table and text, and leads to more noise as full tables always contain redundant information. 
	Secondly, these methods tackle the whole reasoning process as a black box, and lack the interpretability of the intermediate reasoning process. 
	Moreover, the data sparsity problem is also severe, as the high-quality annotated reasoning process is hard to be obtained.
	
	To tackle these challenges, we propose a ChAin-centric Reasoning and Pre-training framework (CARP), which models the intermediate reasoning process across table and text with a hybrid chain for question answering. 
	CARP first formulates a heterogeneous graph, whose nodes are information pieces in the relevant table and passages, to represent the interaction residing in hybrid knowledge. 
	Then, it identifies the most plausible reasoning path leading to the answer with a Transformer-based extraction model. 
	Moreover, to augment the pre-trained model with ability to identify the reasoning process, we propose a novel chain-centric pre-training method, which takes the advantage of the clear table structure and table-passage connections to construct large-scale pseudo reasoning paths, and reversely generate questions. 
	CARP framework has following advantages. 
	Firstly, the hybrid chain models the interaction between table and text, and reduces the redundant information.
	Secondly, it provides a guidance for QA, and better interpretability of the intermediate reasoning process.
	Lastly, both the training of the extraction model and the pre-training corpus construction require no human-annotated reasoning process, which alleviates the data sparsity problem and broadens the potential applications of the framework. 
	
	Experiments show that our system achieves the state-of-the-art result on a large-scale table-and-text open-domain question answering benchmark OTT-QA. 
	Notably, the effectiveness of the chain-centric pre-training method is proved by the significant performance boost of the chain extraction model.
	Results show that incorporating the hybrid chain enhances the QA model, especially for the questions requiring more complicated reasoning process.
	We summarize following contributions:
	\begin{itemize}
		\item[1)] We propose to model the intermediate reasoning process for question answering over table and text, with a fine-grained hybrid chain.
		\item [2)] We propose a novel pre-training method, which captures the reasoning process by pre-training on a synthesized reasoning corpus consisting of large-scale cross-modality reasoning paths and corresponding questions. 
		\item [3)] Experiments show that our system achieves the state-of-the-art result and further analysis proves the effectiveness of utilizing the hybrid chain and the pre-training method.
	\end{itemize}
	\section{Task Definition}
	In this paper, we study the task of question answering over table and text in a challenging open-domain setting, because the supported knowledge is not always provided in a realistic application.
	The task \cite{chen2020open} takes a question as the input, then requires the systems to first retrieve supported tables and passages, and then make inference over the retrieved knowledge to derive a free-formed answer as the output.
	The answer is a span from either the table cells or the passages. 
	One of the core challenges of this task is that problem solving always requires complex reasoning process across table and text, considering the cross-modality interaction and table structure.
	\section{Framework: CARP}
	\begin{figure*}[thbp]
		\centering
		\includegraphics[width=0.92\textwidth]{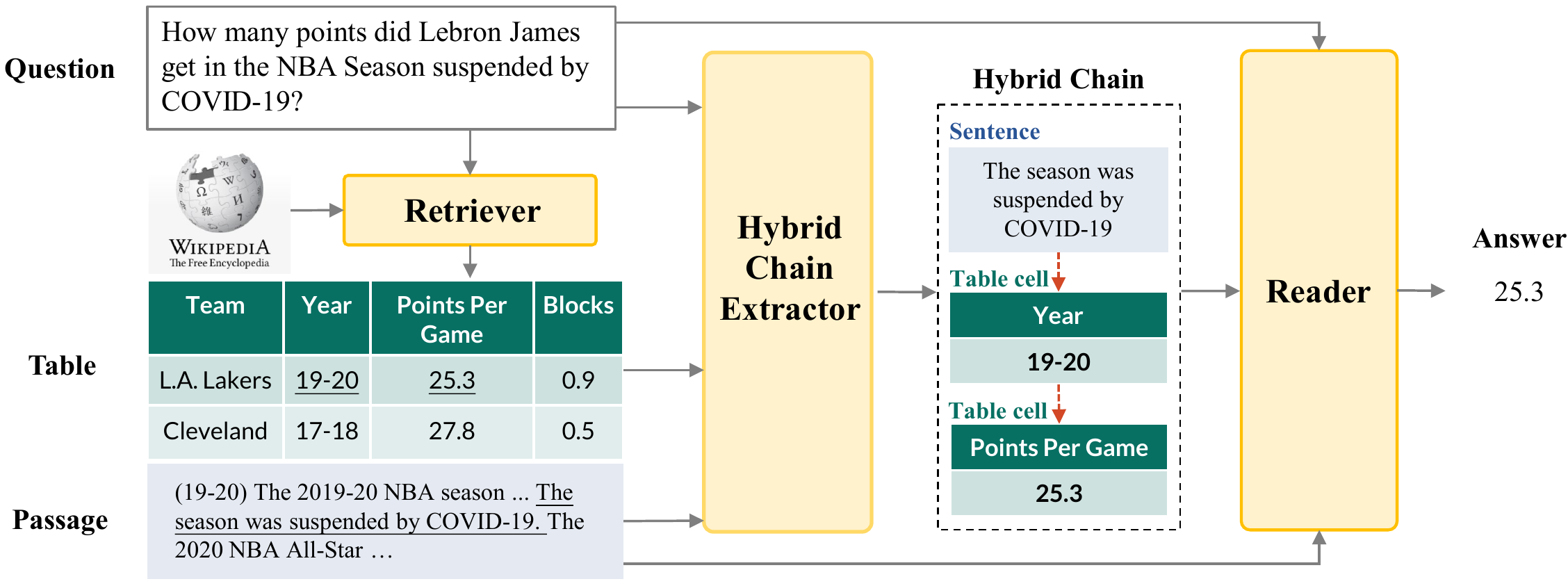}
		\caption{Overview of our system. Retriever (\cref{sec:retriever}) first retrieves knowledge from the corpus for the question. 
		Secondly, hybrid chain extractor (\cref{sec:ec-extractor}) extracts hybrid chains from the knowledge, which is improved by pre-training (\cref{sec:ec-pretrain}). 
		Finally, reader (\cref{sec:qa}) answers the questions with retrieved evidence and extracted hybrid chain. }
		\label{fig:pipeline}
	\end{figure*}
	Fig. \ref{fig:pipeline} shows the pipeline of our CARP framework, which has three main parts: 
	(1) a \textbf{retriever} that retrieves tabular and textual knowledge with the given question (\cref{sec:retriever});
	(2) a \textbf{chain extractor} that extracts hybrid chain from the retrieved knowledge (\cref{sec:ec-extractor}).
	(3) a \textbf{reader} that answers questions with retrieved knowledge and the extracted hybrid chains (\cref{sec:qa}).
	We detailedly illustrate the hybrid chain (i.e., definition, extraction, pre-training, and application in QA), and briefly introduce the retriever.
	\subsection{Hybrid Chain Notation}
	Hybrid chain logically reveals the fine-grained reasoning process from question to the answer across table and text.
	We define the \textbf{hybrid chain} as a sequence of nodes extracted from a fine-grained heterogeneous graph $\mathcal{G}$ , whose nodes $V$ contain the question, cells in the table and sentences in the related passages. 
	One example of the hybrid chain is shown in Fig. \ref{fig:example}. 
	Two nodes in the graph are connected by edges $E$ defined by two types of connections: \textit{structural connections} and \textit{contextual connections}. 
	The former indicates that pairs of cells within a same row (e.g., edge $c$ in Fig. \ref{fig:example}), or a cell to the a sentence in its linked passage (e.g., edge $b$), are structurally connected. 
	The latter indicates that pairs of nodes with relevant context (i.e., entity/ keyword co-occurrence) are contextually connected (e.g., edge $a$ indicates co-occurred keyword ``COVID-19").
	Specifically, we use off-the-shelf named entity recognition model \cite{Peters2017SemisupervisedST} to extract entities, and extract noun phrase and numerical items as keywords from the node context. Moreover, a table cell and a passage is linked by the entity linker as described in \cref{sec:retriever}.

	\subsection{Hybrid Chain Extraction}\label{sec:ec-extractor}
	Here we introduce how to extract hybrid chains, including the model architecture, training and inference process.
	\subsubsection{Model Architecture}
	We tackle the chain extraction as a semantic matching problem, which selects the best chain from several candidate chains.
	Taking a question and a candidate hybrid chain as the inputs, the model calculates the confidence score of the hybrid chain for answering the question. 
	Each candidate hybrid chain is represented as a flattened sequence of its nodes context. Details and an example are given in the Appendix \ref{appendix:chain-extractor}. 
	We utilize rich contextual representations embodied in pre-trained models like RoBERTa \cite{Liu2019RoBERTaAR} to measure the relevance of a question to every chain candidates. 
	Let's take RoBERTa as an example. The input of the hybrid chain extractor is $input = (\texttt{[CLS]}; q;  \texttt{[SEP]}; c_i)$
	where $q$ and $c_i$ indicate tokenized word-pieces of the question and the flattened $i^{th}$  chain candidate. 
	The $\texttt{[SEP]}$ and $\texttt{[CLS]}$ are speicial symbols. The representation $\bm{h}_{c_i}\in\mathbf{R}^d$ is obtained via extracting the hidden vector of the $\texttt{[CLS]}$ token. 
	The score $s^{+}_{c_i}$ for raking the candidates is calculated by:
	\begin{equation}
		(s^{-}_{c_i},s^{+}_{c_i}) = \softmax(\bm{W}\bm{h}_{c_i}+\bm{b})
	\end{equation}
	where $\bm{W}$ and $\bm{b}$ are the learnable parameters.
	The model is trained with the cross-entropy loss.
	\subsubsection{Model Training}
	As mentioned above, the key challenge is constructing the training instances (i.e., ground-truth chains and negative chains), as there is no gold-annotated reasoning process given as a prior. 
	
	We first introduce how to build ground-truth hybrid chains from the heterogeneous graph $\mathcal{G}$ . 
	Partly inspired by \citet{chen2019multi}, we use a heuristic algorithm to derive pseudo ground-truth hybrid chains. 
	Starting from the question, we do the exhaustive search to find all the shortest paths to the nodes containing the answer as the candidate chains. 
	Then, we select the best chain from all the candidate chains that have maximum textual similarity with the question as the final ground-truth hybrid chain, and take it as the positive instance.
%
	To build the hard negative instances, we find the shortest paths from the question node to the non-answer nodes and select the one with maximum textual similarity with the question.
	\subsubsection{Model Inference}
	During Inference, we first build a set of candidate hybrid chains from the graph $\mathcal{G}$, and adopt the extraction model to rank all the chains, and finally select the best chain with highest confidence score.
	
	More specifically, the set of whole candidate hybrid chains contains the shortest paths from the question node to all the other nodes in the graph. 
	Suppose the number of nodes is $n$ in the graph, the number of candidate chains is $\sum^{n-1}_{i=0}SP(i)$, where $SP$ is the number of shortest paths to node $i$.

	\subsection{Chain-centric Pre-training}\label{sec:ec-pretrain}
\begin{figure}[t]
		\centering
		\includegraphics[width=.48\textwidth]{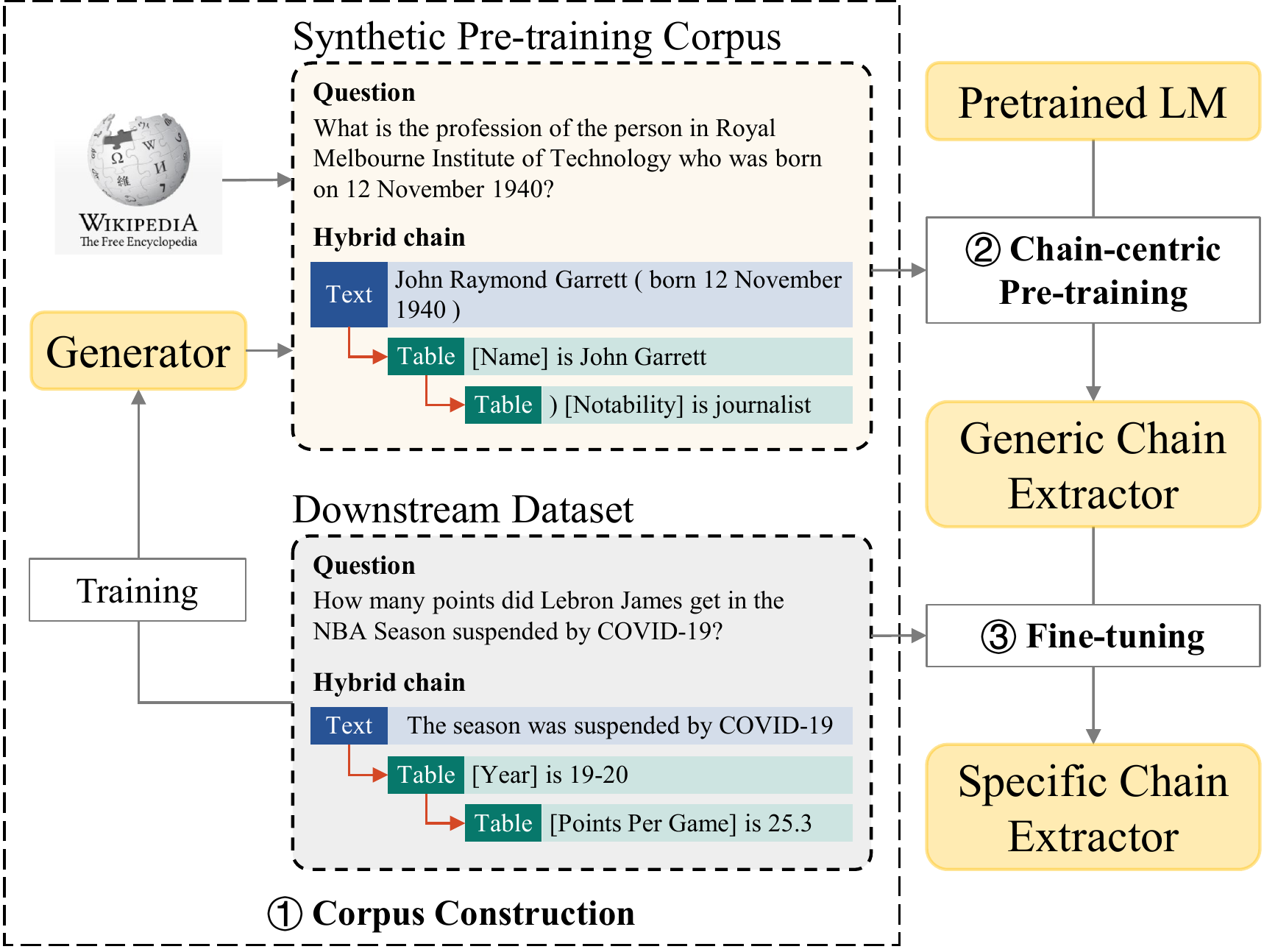}
		\caption{An overview of our pre-training approach. A generic train extractor is first learned by pre-training on the synthesized reasoning corpus. Then, we fine-tune the specific extractor by the downstream data.}
		\label{fig:pretraining}
	\end{figure}

	Pre-training for reasoning is always challenging because high-quality reasoning data is hard to be obtained.
	To better help the pre-trained model in capturing the complicated reasoning process across table and text and alleviate the data sparsity problem, we propose a chain-centric pre-training method.
	The method augments the chain extraction model by pre-training on a synthesized reasoning corpus in larger scale and of higher reasoning complexity.
	The overall process of adopting pre-training strategy is illustrated in Fig. \ref{fig:pretraining}: 
	(1) synthesizing heterogeneous chains from the Wikipedia corpus and reversely generating corresponding questions by a trained generator;
	(2) pre-training a generic extraction model with the synthesized corpus; 
	(3) fine-tuning a specific extraction model with the downstream data. 
	We introduce the pre-training task and the corpus construction.
	
	\subsubsection{Task Formulation}
	The pre-training task can be viewed as a similar semantic matching task that maps hybrid chains to the corresponding pseudo questions. 
	The pre-training objective is in the same spirit of the chain extraction model as described in \cref{sec:ec-extractor}.
	If the model can better distinguish the relevant hybrid chain for answering the given question, then it has deeper understanding of the reasoning process.
	\subsubsection{Corpus Construction}
	
	To construct the large-scale reasoning corpus, we adopt a novel way of first synthesizing heterogeneous reasoning paths, and then reversely generating corresponding questions. 
	Tables in Wikipedia often contain hyperlinks to their related passages. 
	The clear table structure and the explicit table-text links provide natural benefits for automatically synthesizing logically reasonable reasoning paths. 
	
	Therefore, we select semi-structured tables on Wikipedia as the table source, and take the passages hyper-linked to the table cells as the source of passages. 
	The parsed Wikipedia corpus consists of over 200K tables and 3 millions of hyperlinked passages.
	Then, we synthesize pseudo chains with different reasoning depths. 
	For example, to synthesize a 4-hop reasoning path, we randomly select two cells ($c_0, c_1$) within the same row and their related passages ($p_0,p_1$) to form a chain $(p_0, c_0, c_1, p_1)$. Similarly, $(p_0, c_0)$ or $(c_0, c_1,p_1)$ can be selected as a 2-hop or a 3-hop chain, respectively. 
	
	Finally, taking a synthesized flattened chain as the input, we adopt a generation model built based on BART \cite{lewis2019bart} to reversely generate a pseudo question to construct a pair of \textit{(question,  chain)} as a positive instance. 
	It is worth noting that the generation model is trained by the ground-truth \textit{(question,  chain)} pairs as described in \cref{sec:ec-extractor}. 
	To encourage the model to better discriminate relevant chains, we select other chains sampled from the same table with top-$n$ similarity with the question as the hard negative instances.

	\subsection{Hybrid Chain for QA}
	\label{sec:qa}
	Having extracted the hybrid chains for each table segment and its related passages, we need to build a reader model to extract the answer $a$ with the inputs.
	We build a reader model based on a sparse-attention based Transformer architecture \textit{Longformer} \cite{beltagy2020longformer} to process long sequence efficiently.
	With longer limited length up to $4096$ tokens, the reader can read top-$k$ retrieved evidences jointly for question answering.
	The input sequence $x$ is the concatenation of the \textit{question} and top-$k$ pairs of (\textit{table segment, passages}, \textit{hybrid chain}).
	The Longformer encodes the input $x$ of length $T$ into a sequence of hidden vectors:
	\begin{equation}
		\bm{h}(x) =  [\bm{h}(x)_1,\bm{h}(x)_2,\cdots,\bm{h}(x)_T]
	\end{equation}
	The probabilities $p_{start}(i)$ and $p_{end}(i)$ of the start and ending token of the answer $a$ are calculated by:
	\begin{equation}
		\begin{aligned}
			p_{start}(i) = \frac{exp(\bm{W}_{s}\bm{h}(x)_i + \bm{b}_s)}{\sum_{j}exp(\bm{W}_{s}\bm{h}(x)_j + \bm{b}_s)} \\
			p_{end}(i) = \frac{exp(\bm{W}_{e}\bm{h}(x)_i + \bm{b}_e)}{\sum_{j}exp(\bm{W}_{e}\bm{h}(x)_j + \bm{b}_e)}
		\end{aligned}
	\end{equation}
	where $\bm{W}_{s}$, $\bm{W}_{e}$, $\bm{b}_s$, $\bm{b}_e$ are learnable weights and bias parameters of the answer extraction layer. 
	Specifically, to alleviate the bias that the model only looks at the extracted chain, we only set the chain as a guidance of the intermediate reasoning process and force the model to select answer from the tokens of the table and passages.
	\subsection{Knowledge Retrieval}
	\label{sec:retriever}
	Unlike retrievers in text-based open-domain QA systems, the retriever for this task is required to search both supported passages and tables. 
	We briefly introduce the retriever in the last part for integrality, as it is not the main focus of our paper. 
	
	
	

	Instead of independently retrieving tables and passages, we follow \citet{chen2020open} and use an ``early-fusion" mechanism, 
	which groups highly-relevant table cells in a row and their related passages as a self-contained group (\textbf{fused block}). This strategy integrates richer information from two modalities and benefits following retrieval process. 
	We adopt BLINK \cite{wu2019zero} as the entity linker to link a table cell to its related passage. 
	BLINK is a highly effective BERT-based entity linking model and is able to link against all Wikipedia entities. 
	Specifically, taking the cell to be linked and the table metadata as the inputs, BLINK automatically finds the relevant passages for each cell.
	After the linking procedure, we represent each fused block as a row in the table and linked related passages. Further details are given in the Appendix. 
	We then tackle the fused block as a basic unit to be retrieved. 
	
	Finally, a Transformer-based retriever is employed to retrieve top-$k$ fused blocks as the knowledge. We apply a shared RoBERTa-encoder $RoBERTa( \cdot )$ \cite{Liu2019RoBERTaAR} to separately encode questions and fused blocks. 
	The relevance of the question and a fused block is measured by the dot-product over their representations of the $\texttt{[CLS]}$ token.
	We train the retriever model as in \citet{karpukhin2020dense}, where each question is paired 
	with a positive fused block and $m$ negative blocks to approximate the softmax over all blocks. 
	Negative blocks are a combination of in-batch negatives which are fused blocks of the other instances in the mini-batch, 
	and hard negative blocks which are sampled from the other rows in the same table.
	During inference, we apply the trained encoder to all fused blocks and index them with FAISS \cite{Johnson2021Faiss} offline. 
	
	\section{Experiments}
	In this section, we conduct experiments to explore the effectiveness of our method from the following aspects: 
	(1) the performance of our overall system on QA;
	(2) the performance of the hybrid chain extraction model; 
	(3) the ablation study about the pre-training strategy;
	(4) the comprehensive qualitative analysis.
	The retrieval performance and implementation details of all components are described in Appendix \ref{appendix:retriever} and \ref{appendix:detail}, respectively.
	\subsection{Dataset and Evaluation}
	In the real-world scenario, solving many questions requires retrieving supporting heterogeneous knowledge and making reasoning over it.
	Therefore, we evaluate the performance of our approach on the OTT-QA \cite{chen2020open} dataset. 
	OTT-QA is a large-scale table-and-text open-domain question answering benchmark for evaluating open-domain question answering over both tabular and textual knowledge. 
	As the data statistics shown in Table \ref{tab:data-statistic}, OTT-QA has over 40K instances and it also provides a corpus collected from Wikipedia with over 400K tables and 6 million passages. 
	Furthermore, the problem solving in OTT-QA requires complex reasoning steps.
	The reasoning types can be divided into several categories: single hop questions (13\%), two hop questions (57\%), and multi-hop questions (30\%).
	We adopt the exact match (EM) and F1 scores \cite{spider} to evaluate the overall QA performance. 
	\begin{table}[h]
		\centering
		\begin{tabular}{ll}
			\hline
			Type & Numbers \\
			\hline
			Training Examples                               & 41,469  \\ 
			Evaluating Examples  & 2,214          \\
			Testing Examples    & 2,158     \\
			Tables in the Corpus  &  410,740         \\
			Passages in the Corpus & 6,342,314 \\ 
			\hline
		\end{tabular}
		\caption{Data statistics of OTT-QA dataset.}
		\label{tab:data-statistic}
	\end{table}
	
	\subsection{Baselines}
	We compare our system to the following methods:
	\begin{itemize}
		\item \textbf{HYBRIDER} \cite{chen2020hybridqa} is a model that uses BM25 to retrieve relevant tables and passages, and adopts a two stage model to cope with heterogeneous information.
		\item \textbf{Iterative Retriever and Block Reader} The model family is proposed by \citet{chen2020open}, which couples Iterative Retriever (IR) / Fusion Retriever (FR) with Single Block Reader (SBR) / Cross Block Reader (CBR). IR and FR indicate retrieving supported knowledge by standard iterative retrieval or using ``early fusion" strategy to group tables and passages as fused blocks before retrieval, respectively. SBR indicates the standard way of retrieving top-$k$ blocks and then feeding them independently to the reader and selecting the answer with the highest confidence score. CBR means concatenating the top-$k$ blocks together to the reader, with the goal of utilizing the cross-attention mechanism to model their dependency. 
		\item \textbf{DUREPA} \cite{li2021dual} is a recently proposed method that jointly reads tables and passages and selectively decides to directly generate an answer or an executable SQL query to derive the output. 
	\end{itemize}
	
	\subsection{Model Comparison}
	\begin{table}[t]
		\small
		\begin{tabular}{lccccc}
			\hline
			& \multicolumn{2}{c}{Dev} & \multicolumn{2}{c}{Test} \\ \hline
			Models                                    & EM         & F1         & EM          & F1         \\ \hline
			HYBRIDER & 10.3       & 13.0       & 9.7         & 12.8       \\
			IR + SBR  & 7.9        & 11.1       & 9.6         & 13.1       \\
			FR + SBR    & 13.8       & 17.2       & 13.4        & 16.9       \\
			IR + CBR   & 14.4       & 18.5       & 16.9        & 20.9       \\
			FR + CBR     & 28.1       & 32.5       & 27.2        & 31.5       \\
			DUREPA        & 15.8       & --          & --           & --          \\ \hline
			CARP                             & \textbf{33.2}       & \textbf{38.6}       &      \textbf{32.5}  & \textbf{38.5}           \\
			CARP w/o hybrid chain   & 29.4       & 34.2 & -- & -- \\ \hline
		\end{tabular}
		\caption{Performance of different methods on the dev set and the blind test set on OTT-QA. The performance of CARP without hybrid chain is also reported.}
		\label{tab:overall-performance}
	\end{table}
	Table \ref{tab:overall-performance} reports the performance of our model and baselines on the development set and blind test set on OTT-QA. 
	In terms of both EM and F1, our model significantly outperforms previous systems with 32.5\% EM and 38.5\% F1 on the blind test set, and achieves the state-of-the-art performance on the OTT-QA dataset. 
	It is worth noting that, our approach, which exploits explicit hybrid chain, helps the model to capture the reasoning process and boost the performance of the QA model. 
	
	
	\subsection{Evaluation of Chain-centric Reasoning}
	To verify the effectiveness of our proposed hybrid chain, we firstly eliminate hybrid chain from the QA model inputs, and report the result of ``\textit{CARP w/o hybrid chain}" on the development set in Table \ref{tab:overall-performance}.
	 Incorporating hybrid chain into the QA model improves the performance significantly. 
	
	Then, we explore the performance of various variants in hybrid chain extraction, whose backbone is the pre-trained model RoBERTa \cite{Liu2019RoBERTaAR}.
	The variants consider three aspects: (1) encoding strategies; (2) ways of heterogeneous graph construction; (3) negative sampling strategies.
	\begin{itemize}
		\item[(1)] \textbf{Dual Ranking vs Cross Matching}: Dual-tower ranking model \cite{karpukhin2020dense} encodes the question and the hybrid chain separately, and uses the cosine-distance to measure their relevance for ranking. Cross matching means that we use a semantic matching model as described in \cref{sec:ec-extractor}.
		\item[(2)] \textbf{Simple (S) vs Weighted (W)}: Simple indicates the edges in the graph are unweighted. Weighted graph means that the edges connecting highly-related (higher ratio of overlapped keywords) nodes have lower weight, and thus the paths with higher overall relatedness (shorter length) are ranked higher in the ground-truth chain construction (\cref{sec:ec-extractor}). 
		\item[(3)] \textbf{BMNeg vs InnerNeg}: BMNeg means that the most similar chain from other positive instances with BM25 are selected as the negative instance.
		InnerNeg indicates that we select negative instances from other chains constructed from the same fused block, as described in \cref{sec:ec-extractor}.
	\end{itemize}
	\begin{table}[h]
	\small
		\centering
		\begin{tabular}{lll}
			\hline
			Methods                               & Rec@$1$ & Rec@$2$ \\ \hline
			Dual Ranking (W + InnerNeg)  & 61.61      &  73.15          \\
			Cross Matching (W + BMNeg)    & 44.21      & 61.14      \\
			Cross Matching (S + InnerNeg)   & 68.32      & 79.87      \\
			Cross Matching (W + InnerNeg) & 70.75      & 80.19      \\ \hline
		\end{tabular}
		\caption{Performance of the hybrid chain extraction model with different variances.}
		\label{tab:ec-extraction}
	\end{table}
	Table \ref{tab:ec-extraction} reports the performance of the hybrid chain extraction model (without pre-training) with different components. 
	We note that a selected chain is correct when it contains an answer node. 
	We take Recall@$n$ as the evaluation metric.  
	Based on the table, we have following findings.
	Firstly, semantic matching model with cross-attention mechanisms performs better than standard dual-tower ranking model, which verifies that cross-attention mechanism is beneficial for modeling the connections among heterogeneous information.
	Secondly, finding the shortest path in the weighted graph is better than in the simple graph, which shows that modeling the relatedness of nodes is essential in finding a more reasonable hybrid chain.
	Finally, negative sampling strategy is extremely essential for hybrid chain selection. 
	The goal of inference is to select the most plausible chain from several candidate chains sampled from the same fused block. Therefore, sampling hard negative instance from the same fused block is much better than sampling from other training instances.  We take the setting of ``\textit{Cross Matching (W + InnerNeg)}" as the final setting of the extraction model.
	\subsection{Evaluation of Chain-centric Pre-training}
	In this part, we evaluate the effectiveness of the chain-centric pre-training strategy under different settings. 
	The table cells are aligned to the passages according to their hyperlinks in the Wikipedia website. 
	The main variance of pre-training is the different way of constructing instances for training the BART-based generator. \textbf{All} means that we take all the paths from the question node to the answer node as positive chains to train the generator. \textbf{Shortest} indicates that we only select the shortest paths. 
	
	As shown in Table \ref{tab:pre-training}, the pre-training strategy improves the performance of the hybrid chain extraction model by a large margin, showing the effectiveness of chain-centric pre-training in helping the model to capture the intermediate reasoning process with given questions. 
	We believe that several reasons for the improvement of chain-centric pre-training are as follows.
	Automatically synthesizing pre-training data is an effective data augmentation scheme because it can generate data in larger scale and of higher reasoning complexity, which can help the model to better capture the complicated reasoning steps by pre-training. 
	
	Besides, selecting all paths leading to answer as positive chains to train the generator is better than selecting the shortest paths. 
	This observation is intuitively reasonable since the goal of pre-training is to encourage the model to learn a more general reasoning ability with all possible reasoning paths. 
	\begin{table}[h]
	\small
		\centering
		\begin{tabular}{lll}
			\hline
			Methods                               & Rec@1 & Rec@2 \\ \hline
			Extractor    & 70.75      &  80.19  \\
			Extractor + Pre-training (Shortest) & 73.40 & 82.87       \\ 
			Extractor + Pre-training (All) & 74.01 & 83.46       \\ \hline
		\end{tabular}
		\caption{Performance of the chain extraction with chain-centric pre-training under different settings.}
		\label{tab:pre-training}
	\end{table}
	\subsection{Qualitative Analysis}
	
	\pgfplotstableread[row sep=\\,col sep=&]{
    Scale & Baseline & CARP \\
    $1$-hop\,($16\%$)   & 12.5 & 18.8 \\
    $2$-hop\,($60\%$)   & 35.0 & 42.0 \\
    $3$-hop\,($24\%$)   & 33.0 & 46.0 \\
    }\mydata

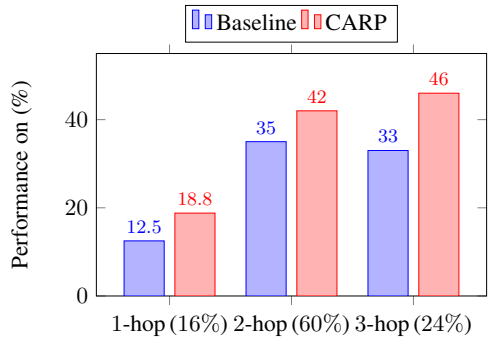
\begin{figure}[t!]
    \centering
    \begin{tikzpicture}[scale=0.8]
        \begin{axis}[
                ybar=5pt,
                bar width=.66cm,
                enlarge x limits=0.3,
                width=.5\textwidth,
                height=.35\textwidth,
                legend style={at={(0.5,1.20)},
                anchor=north,legend columns=-1},
                symbolic x coords={$1$-hop\,($16\%$), $2$-hop\,($60\%$), $3$-hop\,($24\%$)},
                xtick=data,
                nodes near coords,
                every node near coord/.append style={font=\small},
                nodes near coords align={vertical},
                ymin=0,ymax=55,
                ylabel={Performance on  (\%)},
            ]
            \addplot table[x=Scale,y=Baseline]{\mydata};
            \addplot table[x=Scale,y=CARP]{\mydata};
            \legend{Baseline, CARP}
        \end{axis}
    \end{tikzpicture}
    \caption{
    The performance of baseline and our CARP on the randomly selected 100 instances across different hops. 
    The performance on $1$-hop questions is lower mainly because these questions are much less frequent in the dataset \cite{chen2020open}, and always require more complex numerical table understanding.}
    \label{fig:hop-analysis}
    \vspace{-5mm}
\end{figure}

	We randomly select 100 instances from the development set and manually annotate the plausible hybrid chains and conduct qualitative analyses on several aspects: 
	(1) the performance on the questions requiring different reasoning steps;
	(2) a case study by giving an example;
	(3) an analysis of common error types to shed a light on future directions. 
	\paragraph{Performance on M-hop Questions}

	As shown in Fig. \ref{fig:hop-analysis}, we report the performance of the baseline (CARP without hybrid chain) and CARP on the selected questions with different reasoning steps.
	It can be observed that as the number of reasoning steps
	increases, the improvement brought by our method to the baseline becomes more significant. 
	This observation verifies that, the hybrid chain is essential in helping the model to identify the intermediate reasoning steps towards the answer especially when the reasoning is more complicated. 
	Our synthesized pre-training corpus includes higher ratio of 3-hop questions, which enhance the multi-hop reasoning ability of the system.
	
\begin{figure}[t]
		\centering
		\includegraphics[width=.45\textwidth]{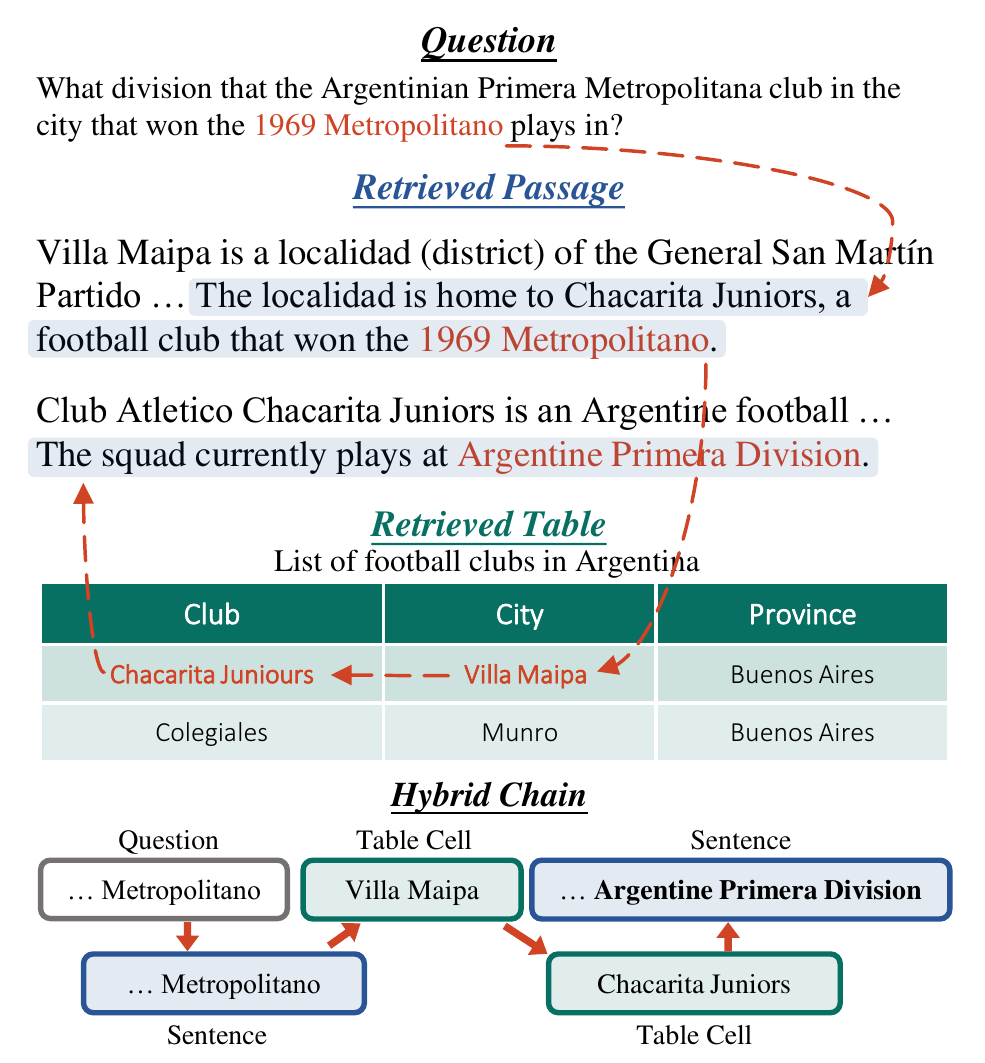}
		\caption{A case study of our approach. The answer is \textit{Argentine Primera Division}. We omit some unimportant sentences in the passage for simplification.}
		\label{fig:case-study}
	\end{figure}

	\paragraph{Case Study} 
	We conduct a case study by giving an example shown in Fig. \ref{fig:case-study}. 
	From the example, our chain extraction model selects a semantic-consistent hybrid chain from the fused block and the QA model correctly predicts the answer with the help of the hybrid chain. 
	This observation reflects that our model has the ability to extract intermediate reasoning process from the given inputs and utilize these information to facilitate the question answering process.
	Hybrid chain also makes the predictions become more interpretable.
	
	\paragraph{Error Analysis}
	We summarize major types of errors to shed a light on future directions. 
	The most common type of errors is caused by the disturbance of  wrongly retrieved fused blocks because we feed top-$k$ fused blocks jointly to the model.
	We observe that although our model finds the correct blocks and identifies correct chains, but the answer is selected from the other blocks. 
	The second type of errors is caused by failing to understand complicated numerical relation when building the chain (e.g.,  ``\textit{finding the $9^{th}$ team}" needs to numerically compare the rank of several teams). 
	Further research can focus on the confidence of the retrieved blocks and the numerical understanding of the table.
	
	\section{Related Work}
	Semi-structured web table is an essential knowledge source that storing
 significant amount of real-world knowledge. 
	Furthermore, since the compact structured representation of table allows it to represent relational facts like numerical facts and collections of homogeneous entities, so table is a great complement to textual knowledge. 
	There has been a growing interest in QA with both tabular and textual knowledge. 
	HybridQA \cite{chen2020hybridqa} is a close-domain table-and-text question answering dataset with ground-truth knowledge provided. 
	In realistic scenario, the supported knowledge is always required to be retrieved from knowledge corpus. 
	There are also other table-based datasets, like WikiTableQuestions \cite{wikitablequestion}, WikiSQL \cite{wikisql}, SPIDER \cite{spider}, and TABFACT \cite{chen2019tabfact}, etc.
	These datasets mainly focus on reasoning on table and may discard some important information stored in textual corpus. 
	We study OTT-QA \cite{chen2020open}, which is a large open-domain table-and-text QA dataset requiring aggregating information from hybrid knowledge. 

	There exist text-based question answering datasets designed in open-domain \cite{joshi2017triviaqa,dunn2017searchqa,nqopen} or multi-hop \cite{yang2018hotpotqa,wikihop} settings.
	Graph-based models \cite{de2018question,fang2019hierarchical, ding2019cognitive} utilize graph structure and graph neural network to model the connections among sentences or entities for multi-hop QA. 
	There are works adopting chain-like reasoning to solve multi-hop textual QA \cite{chen2019multi, asai2019learning, feng2020recoverpath}.
	
	Our approach differs from previous methods mainly in two aspects: 
	(1) our method formulate heterogeneous chain to model the complex reasoning process across table and text; 
	(2) the chain-centric pre-training method can enhance reasoning ability of models by pre-training on a synthesized reasoning corpus, containing heterogeneous reasoning paths and pseudo multi-hop questions. 

	\section{Conclusion}
	In this paper, we present a chain-centric reasoning and pre-training (CARP) framework for table-and-text question answering. 
	When answering the questions given retrieved table and passages, CARP first extracts explicit hybrid chain to reveal the intermediate reasoning process leading to the answer across table and text.
	The hybrid chain provides a guidance for QA, and explanation of the intermediate reasoning process. 
	To enhance the extraction model with better reasoning ability and alleviate data sparsity problem, we design a novel chain-centric pre-training method.
	This method synthesizes the reasoning corpus in a larger scale and of higher reasoning complexity, which is achieved by automatically synthesizing heterogeneous reasoning paths from tables and passages in Wikipedia and reversely generating multi-hop questions.
	We find that the pre-training task boosts performance on the hybrid chain extraction model, especially for questions requiring more complex reasoning, which leads to significant improvement on the performance of the QA model. 
	The hybrid chain also provides better interpretability of the reasoning process.
	Our system achieves the state-of-the-art result on a table-and-text open-domain QA benchmark.
	
	\bibliography{anthology,custom}
	\bibliographystyle{acl_natbib}
	
	\appendix
	\section{Evaluation of Retrieval Model}
	\label{appendix:retriever}
In this part, we evaluate the  retrieval performance of retrievers.
\subsection{Settings}  Our retriever is evaluated on the OTT-QA dataset \cite{chen2020open}, which is a large-scale open-domain question answering dataset over table and text. 
We compare our retriever with the following retrieval methods. (1) BM25 \cite{chen2020open} is a sparse method to retrieve tabular evidence with BM25. It represent the table with the flattened sequence of table metadata (i.e., table title and section title) and table content. (2) Bi-Encoder \cite{Kostic2021MultimodalRO} is a dense retriever  which  uses  a  BERT-based  encoder  for  questions, and a shared BERT-based encoder to separately en-code tables and text as representations for retrieval. (3) TriEncoder \cite{Kostic2021MultimodalRO} is a dense retriever that uses three individual BERT-based en-coder to separately encode questions, tables and text as representations.

\subsection{Evaluation Metrics}
In this experiment, we use two metrics to evaluate the retriever: table recall and fused block recall. 
Table recall indicates whether the top-$k$ retrieved blocks come from the ground-truth table, which is also used in other papers.
However, in table-text retrieval, table recall is imperfect as an coarse-grained metric since our basic retrieval unit is a table-text block.
Therefore we use a more fine-grained and challenging  metric:  fused  block  recall  at top-$k$ ranks, where a fused block is considered a correct match when it meets two requirements: coming from the ground truth table and containing the correct answer. 

\subsection{Performance}
The results are shown in Table \ref{tab:retrieval-overall}. We can find that our retriever substantially outperforms sparse BM25 method and achieves comparable performance with Bi-Encoder and Tri-Encoder.


\begin{table}[htbp]
\small
  \centering
  \resizebox{0.475\textwidth}{!}{
    \begin{tabular}{l|ccc|ccc}
    \hline
    & \multicolumn{3}{c|}{Table Recall} & \multicolumn{3}{c}{Block Recall} \\
    Models & R@1  & R@10 & R@100 & R@1  & R@10 & R@100 \\
    \hline
    BM25 & 41.0  & 68.5   & -  & -  & -  & - \\
    Bi-Encoder & -     & 72.9    & 89.4 & -  & -  & -  \\
    Tri-Encoder & -     & 73.8   & 90.1 & -  & -  & -  \\
    \hline
    Ours & 49.0  &  74.0   & 88.6   &  16.3 & 46.7  & 75.5 \\
    \hline
    \end{tabular}
    }
  \caption{Overall retrieval results on OTT-QA dev set. Table recalls and fused block recalls are reported.}
  \label{tab:retrieval-overall}
\end{table}%

\section{Implementation Details}
\label{appendix:detail}
\subsection{Fused Block Representation}
In this part, we describe how we represent a fused block with a table row and its related passages. Similar to \citet{chen2020open}, we represent each fused block as the concatenation of the table meta data, the cells in the rows, and related passages: 
$Fused\ Block = (\texttt{[TAB]}\ \texttt{[TITLE]}\ title\ 
\texttt{[DATA]}\\ row \texttt{[PASSAGES]}\ passages)$, where $row$ and $passages$ indicate the flattened row and all the related passages of this row, and there is a \texttt{[SEP]} token between cells or passages.
\subsection{Retrieval Model}
In this part, we describe the details of the fused block retrieval model. Our retrieval model follows a typical dual-encoder architecture, which uses a dense encoder $E(\cdot)$ to map any fused block to a $d$-dimensional dense vector and build an index for all the blocks for retrieval.
At query time, the input question $q$ is mapped to a $d$-dimensional dense vector by the same neural encoder $E(\cdot)$, and returns top-$k$ fused blocks that are closest to the question representation.
The similarity of $q$ and $b$ is measured by a dot-product of two vectors:
\begin{equation}
    sim(q, b) =  E(q)^{\top} \cdot E(b).
\end{equation}

In practice, we use a pre-trained RoBERTa-base \cite{Liu2019RoBERTaAR} to initialize our encoder and take the representation at the first token (i.e. \texttt{[CLS]} token) as the the output. At inference time, we apply FAISS \cite{Johnson2021Faiss} to index the dense representations of all fused blocks.
\subsubsection{Training} The training objective aims to maximize the probability of positive pairs. Formally, given a question $q_i$ together with  its positive block $b_i^{+}$ and $m$ negative blocks $\{b_{i,1}^-, ..., b_{i,m}^-\}$, we optimize the loss function as the negative log-likelihood of positive block:
\begin{equation}
\begin{aligned}
    & \qquad L(q_i, b_i^{+}, \{b_{i,1}^-, ..., b_{i,m}^-\}) \\ 
    & = - \log \frac{e^{sim(q_i, b_i^+)}}{ e^{sim(q_i, b_i^+)} + \sum^{m}_{j=1} e^{sim(q_i, b_{i,j}^-)}}.
\end{aligned}
\end{equation}
Following \citet{karpukhin2020dense}, we use $1$ hard negative fused block randomly sampled from the same table, and $m-1$ in-batch negatives during training.
\subsection{Hybrid Chain Extraction Model}
\label{appendix:chain-extractor}
In this part, we describe the example of the flattened hybrid chain and training details of our
hybrid chain extraction model. 
\paragraph{Verbalization of the hybrid chain}
We introduce how to represent hybrid chain with natural language, and enable the powerful pre-trained language model to calculate its contextual representations.
Each node is either the question, a table cell or a sentence in the passages. Therefore, we represent the content in different types of nodes as: ``\textit{[Question] $(\text{question})$}", 
``\textit{[Table] ($\text{column}\_\text{name}$) is ($\text{cell}\_\text{content}$)}" or ``\textit{[Passage]} ($\textit{sentence}$)", respectively. 
\textit{[Question], [Table], [Passage]} denote special symbols. 
Then, we concatenate the context in all the nodes corresponding to their types, and separate them with a ``\textit{[SEP]}" special symbol. In our experiment, we omit the question node from the final sequence, to avoid exceeding the maximum sequence length limit of the pre-trained models.

For example, the hybrid chain in Fig. \ref{fig:example} can be represented as:
``\textit{[Question] How many ... COVID 19? [SEP] [Passage] The season ... COVID-19. [SEP] [Table] Year is 19-20. [SEP] [Table] Points is 25.3.}"
 
\paragraph{Training Details}
We employ cross-entropy loss as the
loss function. We apply AdamW as the optimizer
for model training. We employ RoBERTa-Base as
the backbone of our approach. We set the learning rate as 1e-5, warmup
step as 0, batch size as 16 per GPU, and set max sequence
length as 512. The training time for one
epoch takes 1 hours on 8 V100 GPUs.

\subsection{Chain-centric Pre-training}

\paragraph{Corpus Construction}

When constructing the pre-training corpus, we use 3 millions pairs of \textit{(question, hybrid chain)} as the positive training instances, and search for the same number of hard negative instances, and the final pre-training corpus contains nearly 6 millions of training instances. 
It worth noted that, to avoid the bias caused by the length of the hybrid chain, we automatically synthesize hybrid chains with different length various from 1 to 4. 
The ratio of the synthesized chains with different lengths are: 1-hop (0.1); 2-hop (0.25); 3-hop (0.35); 4-hop (0.3). 
As for the pseudo questions generator, we employ BART-Large as the backbone.
It is firstly trained upon pairs of our extracted hybrid chains and questions from the OTT-QA dataset.
During training, its learning rate is set as 3e-5, warmup step is as 2000, and batch size is as 8 per GPU.
The training time for one epoch takes nearly 2 hours on 8 V100 GPUs.

\paragraph{Training Details}

Then we describe the training details of the chain-centric pre-training. 
Similar to the implementation details of hybrid chain extractor, we employ cross-entropy loss as the
loss function. 
We adopt \textit{RoBERTa-Base} \cite{Liu2019RoBERTaAR} as the model backbone and use AdamW as the optimizer for model training
the backbone of our approach. We set the learning rate as 3e-5, warmup
step as 0, batch size as 32 per GPU, and set max sequence
length as 512. The training time for one
epoch takes 8 hours on 8 V100 GPUs.
\subsection{QA Model}
We employ the \textit{Longformer-Base} \cite{beltagy2020longformer} as the backbone of our QA model. We set batch size as 2 per GPU, set max sequence length as 512, and set document stride as 3072. The learning rate is 1e-5. The training time for one
epoch takes 3 hours on 8 V100 GPUs.
We concatenate top-15 fused block as the evidence for both training and inference. We adopt AdamW as the optimizer, and use cross entropy as the loss function. During training and inference, we force the model to only select the answer from the tokens of the fused blocks.

\end{document}